\documentclass[runningheads]{llncs}
\usepackage[T1]{fontenc}
\usepackage{graphicx}

\usepackage{booktabs}
\usepackage{CJKutf8}
\usepackage[misc]{ifsym}
\newcommand{\corr}{\textsuperscript{(\Letter)}}
\usepackage{multirow}
\usepackage{fontawesome} 
\usepackage{mwe}
\usepackage{cite}
\usepackage{subfigure}
\usepackage{extarrows}
\usepackage{hyperref}
\usepackage{url}
\usepackage{booktabs}
\usepackage{ulem}
\usepackage{bm}
\usepackage{enumitem}
\usepackage{algorithm}
\usepackage{algorithmic}

\usepackage{amsmath,amssymb,amsfonts}
\usepackage{graphicx}
\usepackage{textcomp}
\usepackage{xcolor}

\usepackage{makecell}

\usepackage{tabularx, array}
\usepackage{orcidlink}
\usepackage[ruled,vlined,algo2e]{algorithm2e}
\begin{document}
\title{Unlocking the Power of Large Language Models for Multi-Table Entity Matching}
\titlerunning{LLM4MEM}

\author{Yingkai Tang\inst{1,2}\orcidlink{0009-0001-3346-5512} \and
Taoyu Su\inst{1,2}{\corr}\orcidlink{0009-0003-1674-7635} \and
Wenyuan Zhang\inst{1,2}\orcidlink{0000-0002-7287-6883} \and
Xiaoyang Guo\inst{3}\orcidlink{0009-0000-3660-5990} \and
Tingwen Liu\inst{1,2}\orcidlink{0000-0002-0750-6923}}

\authorrunning{Y. Tang et al.}
\institute{Institute of Information Engineering, Chinese Academy of Sciences, Beijing, China 
\email{\{tangyingkai,sutaoyu,zhangwenyuan,liugtinweng\}@iie.ac.cn}
\and School of Cyber Security, University of Chinese Academy of Sciences, Beijing, China \and School of Information Engineering China University of Geosciences Beijing, Beijing, China \email{2004230026@cugb.edu.cn}}
\maketitle              
\begin{abstract}

Multi-table entity matching (MEM) addresses the limitations of dual-table approaches by enabling simultaneous identification of equivalent entities across multiple data sources without unique identifiers.
However, existing methods relying on pre-trained language models struggle to handle semantic inconsistencies caused by numerical attribute variations.
Inspired by the powerful language understanding capabilities of large language models (LLMs), we propose a novel LLM-based framework for multi-table entity matching, termed LLM4MEM.
Specifically, we first propose a multi-style prompt-enhanced LLM attribute coordination module to address semantic inconsistencies.
Then, to alleviate the matching efficiency problem caused by the surge in the number of entities brought by multiple data sources, we develop a transitive consensus embedding matching module to tackle entity embedding and pre-matching issues.
Finally, to address the issue of noisy entities during the matching process, we introduce a density-aware pruning module to optimize the quality of multi-table entity matching.
We conducted extensive experiments on 6 MEM datasets, and the results show that our model improves by an average of 5.1\% in F1 compared with the baseline model.
Our code is available at {\small\url{https://github.com/Ymeki/LLM4MEM}}.

\end{abstract}
\section{Introduction}
\begin{figure}[htbp]
    \centering
    \includegraphics[width=0.8\linewidth]{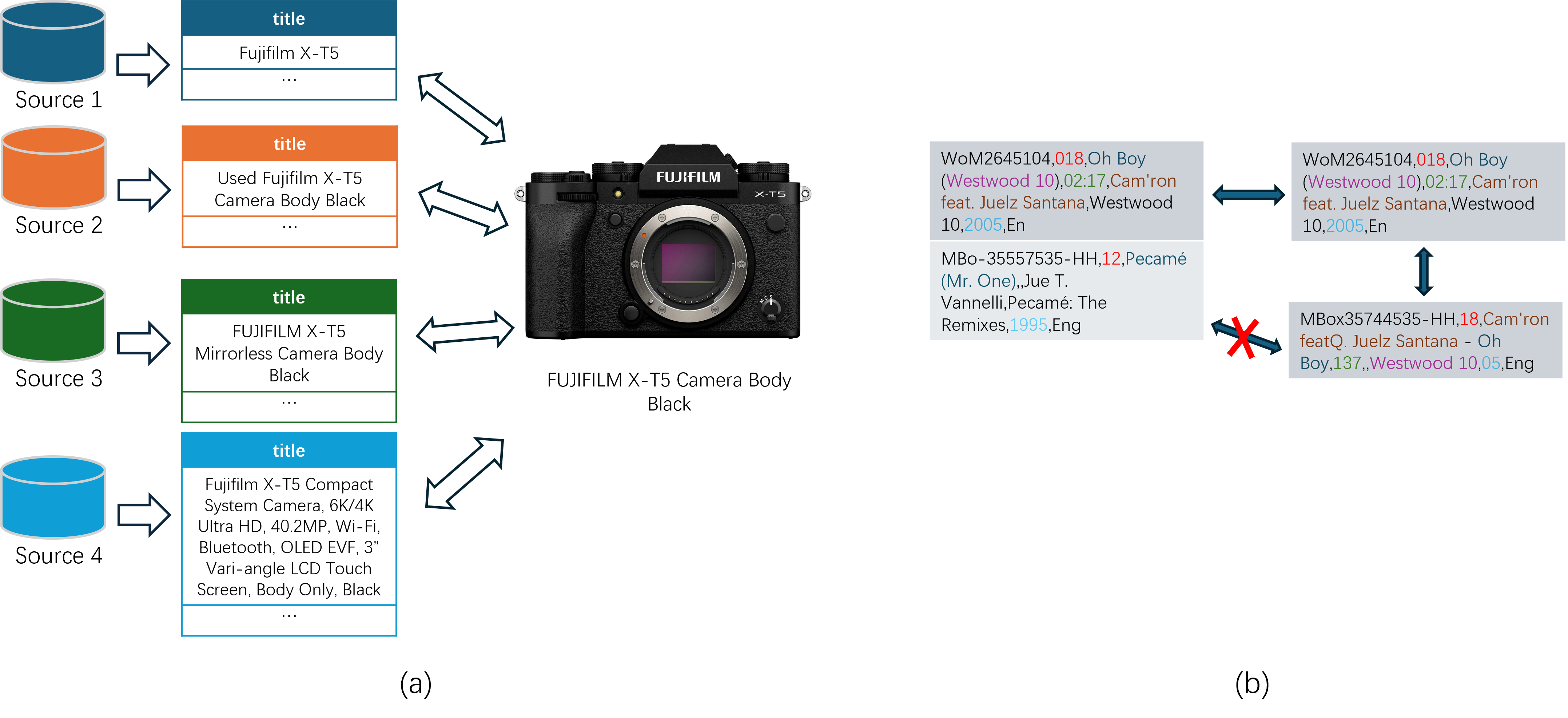}
    \caption{An example of Multi-Table Entity Matching.}
    \label{fig:MEM-example}
\end{figure}

\textit{Entity matching (EM)}, also referred to as record linkage or data deduplication, is a critical task that identifies equivalent entities across diverse data sources in the absence of unique identifiers. 
While existing studies~\cite{ditto,PromptEM,Auto-FuzzyJoin,DADER} focus on dual-table entity matching, their reliance on the assumption that entities originate from only two data tables significantly limits their practical applicability. 
To address this limitation, the \textit{Multi-Table Entity Matching (MEM)} task has been introduced, enabling the simultaneous processing and matching of entities across multiple data tables~\cite{MultiEM,MSCD-HAC}.
As illustrated in Fig.~\ref{fig:MEM-example}, this task is exemplified by three entities from distinct data sources (e.g., \textit{FUJIFILM X-T50 Digital Mirrorless Camera, Silver}), which represent the same real-world entity despite variations in titles, colors, and Memory details.
A key challenge in achieving effective multi-table matching lies in ensuring feature consistency for identical entities across heterogeneous data sources. 
Early approaches~\cite{Auto-FuzzyJoin,MSCD-HAC} relying on n-gram tokenization and string-based similarity metrics fail to capture contextual semantics. 
Although MultiEM~\cite{MultiEM} addresses this limitation by leveraging Sentence-BERT~\cite{Sentence-BERT} for semantic embedding generation, it still struggles with numerical discrepancies (e.g., 64GB vs. 64*1024MB). 
Furthermore, the exponential growth of entity volume in multi-table scenarios poses significant computational challenges, as existing clustering-based~\cite{MEM-KEOD,MEM-BTW} and dual-table extension~\cite{Auto-FuzzyJoin} methods exhibit notable efficiency bottlenecks.

\setlength{\abovecaptionskip}{2pt} 
\setlength{\belowcaptionskip}{0pt} 
\setlength{\intextsep}{12pt}  
\setlength{\textfloatsep}{12pt}
Leveraging the advanced language understanding capabilities of large language models (LLMs), we propose LLM4MEM, a novel framework for multi-table entity matching.
Our approach proposes a multi-style prompt-enhanced LLM attribute coordination module to address semantic inconsistencies. 
Additionally, we develop a transitive consensus embedding matching module to tackle entity embedding and pre-matching issues, complemented by a density-aware pruning module to optimize the efficiency of multi-table entity alignment.
Our contributions can be summarized as follows:
\begin{itemize}
 \item We propose LLM4MEM, an automated entity matching framework leveraging pre-trained large language models to systematically regularize matching data and improve data quality.
 \item By utilizing large language models to understand and generate natural language, we overcome the limitations of traditional methods that rely on manual rules and external knowledge bases, thereby improving the robustness and accuracy of entity matching.
 \item We achieved state-of-the-art performance in experiments on 6 benchmark datasets, demonstrating the superiority of the LLM4MEM method and its potential for practical applications. 
\end{itemize}

\section{Related Work}

\subsection{Entity Matching Techniques and Challenges}
Entity matching (EM) has evolved through three paradigms: early rule-based methods~\cite{GZ2017} used manually crafted similarity metrics, crowdsourcing approaches~\cite{ZB14} leveraged human computation, and machine learning introduced supervised~\cite{ditto} and unsupervised~\cite{Auto-FuzzyJoin} techniques. Deep learning models like DeepMatcher~\cite{DeepMacher} advanced feature modeling, while pretrained language models (PLMs)~\cite{SBERT} enabled semantic-aware matching via transfer learning~\cite{CollaborEM} and prompt engineering~\cite{PromptEM}.  

Traditional EM methods focus on pairwise matching and face challenges in multi-table scenarios due to combinatorial complexity. For example, MSCD-HAC~\cite{MSCD-HAC} struggles with schema heterogeneity, ALMSER-GB~\cite{ALMSER-GB} incurs quadratic comparisons, and MultiEM~\cite{MultiEM} relies heavily on precise schema alignment. Common limitations include sensitivity to schema drift, exponential complexity growth, and reliance on structured attribute correspondences.

\subsection{LLM-enhanced Matching Paradigms}
The emergence of large language models (LLMs) has transformed EM. Initial explorations~\cite{Match-Compare-or-Select} treated EM as text pair classification using LLM prompting, while Batcher~\cite{CE2024} introduced cost-effective batching strategies. Recent work~\cite{Match-Compare-or-Select} demonstrated LLMs' zero-shot matching potential through innovative prompts. However, existing LLM-based methods focus on pairwise matching and lack mechanisms for handling multi-table correlations and schema inconsistencies.

Our LLM4MEM framework uniquely addresses these gaps by employing dynamic schema alignment via multi-prompt attribute coordination, achieving linear complexity through transitive consensus propagation, and enabling density-aware error correction for cross-table inconsistencies. 
Unlike prior LLM applications that directly classify entity pairs, we leverage LLMs as semantic regularizers and cross-source consensus builders, ensuring robust multi-table alignment without labeled data.

\section{Task Formulation}

\textbf{Multi-table Entity Matching (MEM)} aims to identify all records from multiple structured tables that refer to the same real-world entity. 
Given a collection of $n$ tables $\mathcal{E} = \{E_1, E_2, \dots, E_n\}$,
where each table $E_{i}$ contains $m_{i}$ records, denoted as $E_i = \{r_{i}^{(1)}, r_{i}^{(2)}, \dots, r_{i}^{(m_{i})}\}$, where $r_{i}^{(j)}$ denotes the $j$-th record in the $i$-th table.
The core objective is to identify all cross-table record tuples $\{( r_{1}^{(j_1)}, \ldots, r_{n}^{(j_n)} )\} $ that correspond to the same real-world entity \( T_k \in \mathcal{T} \), where \( \mathcal{T} \)  represents the set of real-world entities that covered across all tables.

Multi-table entity matching consists of two steps: merging and pruning.
The merging step identifies potential matching tuples across tables, while the pruning step selects the most accurate matches.
Unlike dual-table EM, which focuses on finding matching entity pairs, multi-table EM identifies matching tuples representing equivalent entities across multiple tables.

\section{Method}

\begin{figure}[t]
\includegraphics[width=\textwidth]{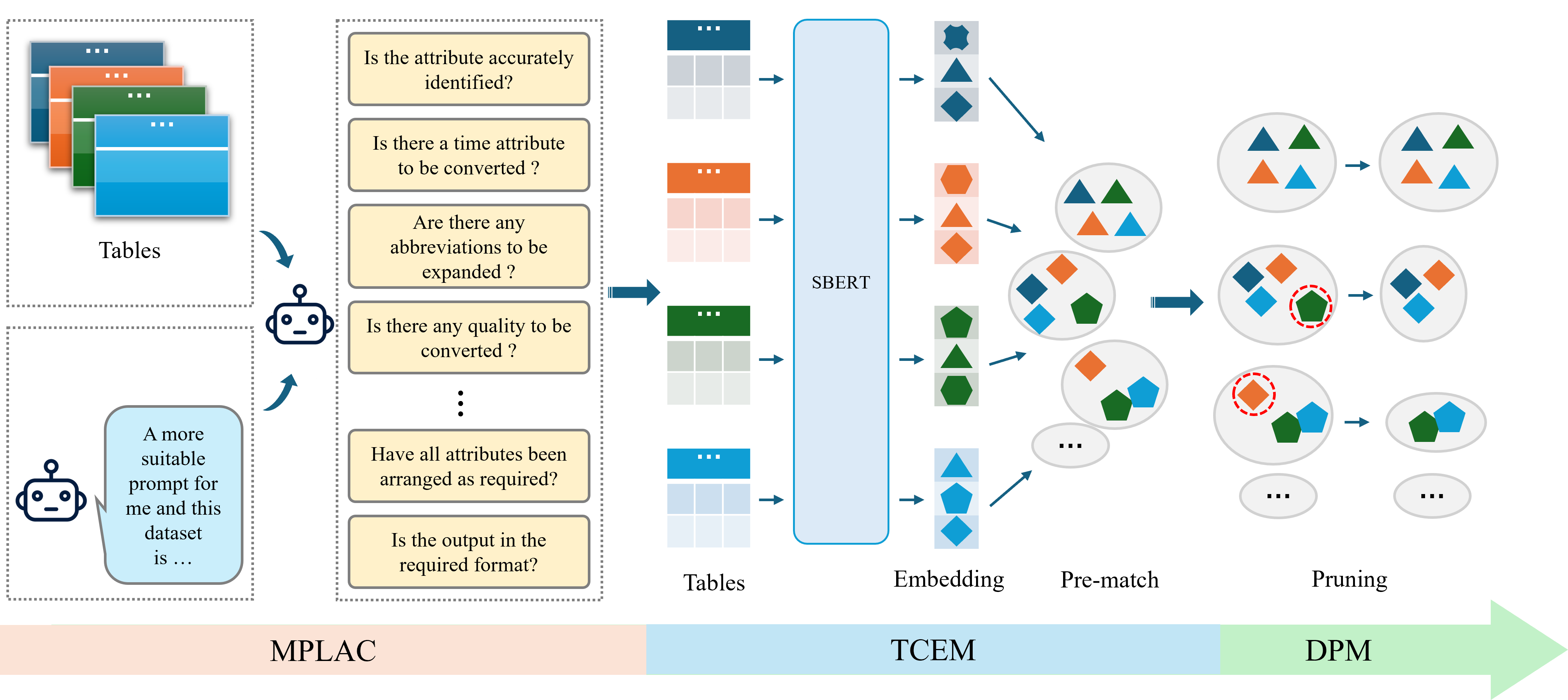}
\caption{The overview of LLM4MEM framework.} \label{fig-overview}
\end{figure}

Our LLM4MEM framework, shown in Figure~\ref{fig-overview}, consists of three modules: the Multi-style Prompt-enhanced LLM Attribute Coordination module to address chaotic and regularized data attributes, the Transitive Consensus Embedding Matching module for entity embedding and pre-matching, and the Density-aware Pruning module to refine final matches. 
Section 4.1 introduces the attribute coordination module, Section 4.2 details embedding matching optimization, and Section 4.3 explains density-aware pruning.

\subsection{Multi-style Prompt-enhanced LLM Attribute Coordination Module}

To address entity attribute confusion that impacts matching quality, we propose the Multi-style Prompt-enhanced LLM Attribute Coordination Module (MPLAC), which leverages the text understanding ability of LLMs to organize and format data for output. 
We first extract a subset of data samples from the dataset, denoted as $\mathcal{D}_{sample}$, to help the LLM understand the format of the samples to be processed.
In the prompt, we explicitly instruct the LLM to focus on and normalize specific types of data (e.g., numerical values and time), including standardizing time formats and numerical units, as demonstrated in Table~\ref{tab-prompt}.

\begin{table}[h]
\caption{Example of Partial Data Conversion Rules.}
\label{tab-prompt}
\renewcommand\arraystretch{1.0}
 \begin{tabular}{llll}
 \toprule
 {\textbf{Type}} & {\textbf{Turn into }} & {\textbf{Prompt}} & {\textbf{Example}} \\
\midrule
\multirow{3}{*}{time} & \multirow{3}{*}{second} & \multirow{3}{*}{Convert all durations to seconds} & {02:12->137sec,}\\
 & & & {137000->137sec,} \\
 & & & {2.283->137sec} \\
\midrule
\multirow{2}{*}{year} & \multirow{2}{*}{YYYY} & \multirow{2}{5cm}{Convert two-digit years to four-digit years} & 05->2005, \\
 & & & 95->1995 \\
\midrule
abbreviation & \multirow{2}{*}{complrtion} & \multirow{2}{5cm}{Expand and complete abbreviations.} & En/Eng->English, \\
of nouns & & & Fr->French \\
\midrule
\multirow{2}{*}{sort number} & \multirow{2}{*}{ordinal number} & \multirow{2}{5cm}{Convert numeric values to ordinal format} & 01->1st, \\
 &  & & 2->2nd \\
\midrule
weight & unified unit & Unify the weight in units of g & 0.001kg->1g \\
\midrule
capacity & unified unit & Unify the capacity in units of L & 2500ml->2.5L \\
 \bottomrule
 \end{tabular}
\end{table}

\begin{figure}[h]
\includegraphics[width=\textwidth]{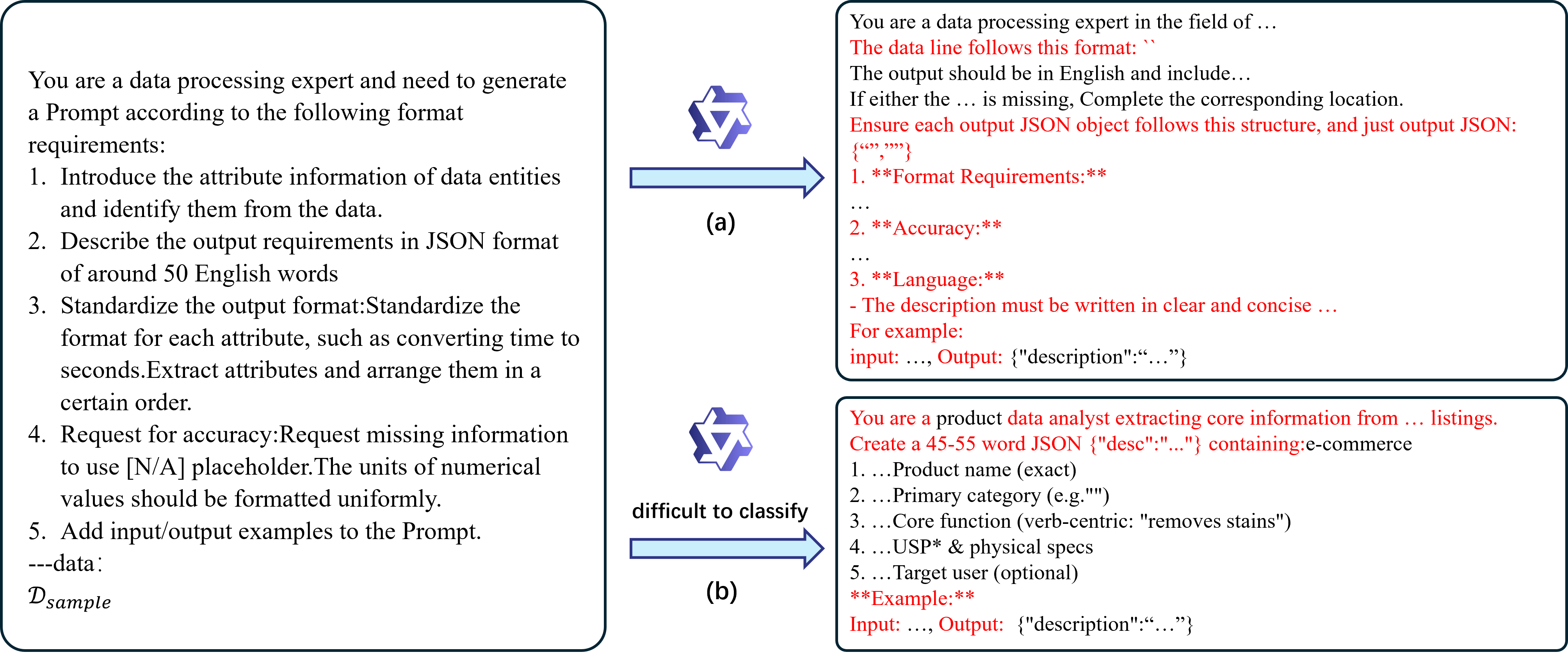}
\caption{The process of generating $Dataset_{AC}$ from $\mathcal{D}_{sample}$ and $Dataset$, with (a) the prompt scheme for a simple dataset and (b) the prompt scheme for a difficult dataset.
}
\label{fig-prompt}
\end{figure}

We integrate our task requirements with $\mathcal{D}_{sample}$ and input them into the corresponding LLM.
After a basic normalization process, we derive a rule-based instruction fine-tuned prompt, denoted as $prompt_{task}(\mathcal{D}_{sample})$. 
For data with unified content attributes, we generate more refined prompts, as shown in Figure~\ref{fig-prompt}.
Each piece of data in the dataset is processed into a string and combined with $prompt_{task}(\mathcal{D}_{sample})$.
The LLM processes each piece of data in sequence, obtaining normalized and enhanced data with attributes, denoted as:
\begin{equation}\label{eq:dataset_ac}
Dataset_{AC} = LLM(prompt_{task}(\mathcal{D}_{sample}),Dataset).
\end{equation}
Here, $Dataset$ represents the complete original dataset, and $Dataset_{AC}$ denotes the attribute-enhanced dataset, which will be used for subsequent processing.
This module aligns dataset entries at the attribute level to enhance the accuracy and robustness of subsequent matching tasks, ultimately producing more precise matching results, as demonstrated in Figure~\ref{fig-prompt}.

\subsection{Transitive Consensus Embedding Matching Module}

To address the challenge of comparing entities across multiple sources in entity matching, we introduce the Transitive Consensus Embedding Matching Module (TCEM).
First, we compute embeddings for entities in all tables $\mathcal{E}$ within the dataset $Dataset_{AC}$. 
Let $h_{e_i} \in \mathbb{R}^d$ denote the embedding vector of the entity $e$ in the $i$-th table, where $d$ is the embedding dimension. 
Next, we use HNSW~\cite{HNSW}, a technique based on approximate nearest neighbor (ANN) search via a navigable small-world graph, to construct indices for each pair of tables.
These indices are utilized to identify the top-1 matching entity pairs, denoted as $\mathcal{P}$, where the cosine similarity between the embeddings is less than a threshold $\lambda$.
The matching entity pairs are defined as follows:
\begin{equation}\label{eq:rank_top}
\mathcal{P} = \{(e_{i},e_{j})\mid e_{i} = \mathrm{Top1}(e_{j}), e_{j} = \mathrm{Top1}(e_{i}), \text{and} \ \mathrm{dist}(h_{e_i} , h_{e_j}) \leq \lambda\},
\end{equation}
where $\mathrm{dist}(\cdot,\cdot)$ represents the distance function.
The $\mathrm{Top1}$ function identifies the highest cosine similarity match between entities in different tables.

To preserve semantic consistency, we construct equivalence classes for the set $\mathcal{P}$ using transitive closures, merging directly or indirectly associated entities into a unified entity. 
The formal definition is as follows:
\begin{equation}\label{eq:G_p}
G_p = \{ e_i, e_j, \ldots, e_n \mid (e_i, e_j) \in \mathcal{P}, \ldots, (e_j, e_n) \in \mathcal{P} \},
\end{equation}
where $G_p$ represents the merged entity set.
Finally, by applying transitivity, all matching entity pairs are merged to form tuples $\mathcal{G} = \cup \{ G_p \}$, which yields pre-matched tuples based on embedding similarity for cross-table entity matching.

\subsection{Density-aware Pruning Module}

The transitive consensus embedding matching generates predicted tuples $\mathcal{G}$ in the final merged table. 
However, because the merging process only considers the locality of two tables, the pre-matched tuples still contain noise. 
For instance, missing entries for an entity may cause incorrect merging of additional entities into the tuple.

To improve matching quality, we introduce the Density-aware Pruning Module (DPM), a pruning algorithm based on spatial density constraints to refine $\mathcal{G}$. 
The process involves defining the density calculation function and categorizing entities into three types:

\textbf{Core Entity}: Let $\mathcal{S}_{e}$ represent the entity set, $d \in \mathbb{R^+}$ be the neighborhood radius threshold, and $\rho_{min} \in \mathbb{N^+}$ be the density threshold. 
For any entity ${e_i} \in \mathcal{S}_{e}$, its neighborhood is defined as:
\begin{equation}\label{eq:core_ent_1}
\mathcal{N}_d(e_i) = \{e_j \in G_p \mid \mathrm{dist}(h_{e_i}, h_{e_j}) \leq d\},
\end{equation}
where $dist(\cdot, \cdot)$ is the normalized semantic distance, and $h_{e_i}, h{e_j}$ represent the embeddings of $e_i$ and $e_j$ respectively.
An entity $e_i$ is considered a core entity if:
\begin{equation}\label{eq:core_ent_1}
e^{ce}_i = \{e_i \in G_p \mid |\mathcal{N}_d(e_i)| \geq \rho_{min}\}.
\end{equation}

\textbf{Reachable Entity}: 
A reachable entity $e_j$ is an entity that does not meet the core entity criteria but has a core entity $e^{ce}$ within its neighborhood.
It is defined as:
\begin{equation}\label{eq:reach_ent}
e^{re}_j = \{e_j \in G_p \mid \exists e^{ce} \in \mathcal{N}_d(e_j), |\mathcal{N}_d(e_j)| < \rho_{min}\}.
\end{equation}

\textbf{Noise Entity}: 
A noise entity $e_k$ is an entity that is neither a core nor a reachable entity. 
It is classified as:
\begin{equation}\label{eq:noise_ent}
e^{ne}_k = \{e_k \in G_p \mid e_k \neq e^{ce}, e_k \neq e^{re}\}.
\end{equation}

Based on these definitions, all entities are processed as follows: core entities are retained for their accurate textual representation, reachable entities are preserved despite potential property variations, and noisy entities with significant semantic deviations are removed from $G_p$.

\section{Experiments}
\subsection{Datasets and Evaluation Metrics}

\paragraph{Datasets.} 
We use six publicly available real-world datasets from diverse domains with varying sizes and sources. 
Dataset details are shown in Table~\ref{tab-datasets}. 
Geo contains geographical entities from four sources (DBpedia, Geonames, Freebase, NYTimes) and was used in the OAEI competition. Music is based on MusicBrainz records, with duplicates generated using DAPO across five sources, where 50\% of original records have duplicates in 2–5 sources with high corruption levels for stress-testing EM.
Shopee is from the 2021 Kaggle Shopee Price Match Guarantee dataset~\cite{shopee-product-matching}, retaining only the title field, while the other five datasets are from~\cite{Saeedi2021MatchingEF}. 
Key attributes include dataset name, domain, entity properties, number of tables (src), entities (entity), and clusters.

\paragraph{Evaluation metrics.}  
Following prior works, we use precision (P), recall (R), and F1 score (F1) as evaluation metrics. 
In multi-table entity matching, a predicted tuple is considered correct only if it perfectly matches the ground truth tuple.

\begin{table}[h]
    \footnotesize
    \centering
    \caption{Statistics of the Datasets Used in Experiments. "Src" refers to the quantity of sources, as shown in $n$ in Chapter 3. "Entity" refers to the number of entity entries contained in the entire data. "Tuples" Indicates the number of matching tuples.}\label{tab-datasets}
    \label{ablation}
    \setlength\tabcolsep{6.0pt}
    \renewcommand\arraystretch{1.6}
    {
    {
    \begin{tabular}{@{}lccccc@{}}
    \toprule
    \multirow{1}{*}{\bf Domain} & \multicolumn{1}{c}{\bf Entity properties} & \multicolumn{1}{c}{\bf Dataset} &  \multicolumn{1}{c}{\bf Src} & \multicolumn{1}{c}{\bf Entity} &  \multicolumn{1}{c}{\bf Tuples}   \\

    \midrule
    geography & tid,city,longitude,latitude & Geo & 4 & 3,054 & 820 \\
 \cline{1-6}
 \multirow{3}{*}{music} & \multirow{3}{*}{\makecell{tid,number,title,length,\\artist,album,year,language}} & Music-20K & 5 & 19,375 & 5,000 \\
 \cline{3-6}
 ~ & ~ & Music-200K & 5 & 193,750 & 50,000 \\
 \cline{3-6}
 ~ & ~ & Music-2M & 5 & 1,937,500 & 500,000 \\
   \cline{1-6}
{person} & \makecell{tid,givenname,surname,\\ suburb,postcode} & {Person} & {5} & {5,000,000} &{500,000}\\
 \cline{1-6}
 product & tid,title & Shopee &  20 & 32,562 & 10,962 \\
\bottomrule
\end{tabular}
}}
\end{table}
\subsection{Baselines}

We compared LLM4MEM with 6 baselines, including dual-table matching methods, a state-of-the-art (SOTA) dual-table approach, and methods designed for multi-table entity matching, including an SOTA multi-table method. 
In the experiments, dual-table methods were adapted using chain matching and evaluated in the multi-table EM setting.
\begin{itemize}
 \item \textbf{PromptEM~\cite{PromptEM}} is a prompt-tuning based approach for low-resource generalized entity matching. 
 \item \textbf{Ditto~\cite{ditto}} is a supervised EM approach that fine-tunes a pretrained language model with labeled data. 
 \item \textbf{AutoFJ~\cite{Auto-FuzzyJoin}} is an unsupervised fuzzy join framework that can be used for dual-table entity matching. 
 \item \textbf{ALMSER-GB~\cite{ALMSER-GB}} is a graph-boosted active learning method for multi-source entity matching. 
 \item \textbf{MSCD-HAC~\cite{MSCD-HAC}} is an extended hierarchical agglomerative clustering algorithm for clustering entities from multiple sources. 
 \item \textbf{MultiEM~\cite{MultiEM}}is an unsupervised multi-table entity matching method that enhances entities and other methods.
\end{itemize}

We adhere to prior research settings. 
For supervised/semi-supervised methods requiring training samples (e.g., PromptEM, Ditto, ALMSER-GB), we randomly sample 5\% of the benchmark data for training and 5\% for validation. 
The test set includes the complete benchmark data, with a comprehensive evaluation conducted for each correctly matched pair and $S_{ns}$ randomly sampled non-matching pairs. For small datasets (Geo, Music-20K, Shopee), $S_{ns}=100$, and for large datasets (Music-200K, Music-2M, Person), $S_{ns}=500$.

\subsection{Implementation Details}
We used LLama3.1-8B~\cite{llama3}, Qwen2.5-7B~\cite{qwen2.5}, and Falcon3-8B~\cite{Falcon3} as core models. Structured prompts and entity info were fed via message-passing. Key params: $n=1$ for unique outputs, $max\_token=64$, $temperature=0.0$ for deterministic decoding, $top\_p=0.95$ for balance. Hardware: NVIDIA A800 GPUs with VLLM~\cite{vllm}, using continuous batch scheduling and page attention for efficiency.  

SentenceBERT~\cite{SBERT} was selected for embeddings. Params: $max\_seq\_length=64$, $batch\_size=512$, $L_2$ norm ($normalize\_embeddings=True$) to constrain vectors, improving cosine similarity robustness. Hyperparameter $\lambda$ experiments in the analysis section.  

Pre-matched tuples from TCEM are similarity-filtered with limited entities. Minimum retention threshold $\rho_{min}=2$ balances denoising and info integrity. Neighborhood radius $d$ experiments are detailed in the analysis section.

\subsection{Main Results}

\begin{table}[!t]
    \footnotesize
    \centering
    \caption{Matching Performance of All the Methods. Best F1 in \textbf{bold}, second best in \underline{underline}. The symbol “--” denotes that the method can NOT produce any result after 7 days in our experimental settings. \label{tab-f1-1}}
    \label{ablation}
    \setlength\tabcolsep{6.0pt}
    \renewcommand\arraystretch{0.7}{
    {
    \begin{tabular}{@{}lccccccccc@{}}
    \toprule
    \multirow{2.5}{*}{\bf Model} & \multicolumn{3}{c}{\bf Geo} & \multicolumn{3}{c}{\bf Music-20K} &  \multicolumn{3}{c}{\bf Music-200K}    \\
    \cmidrule(r){2-4}\cmidrule(r){5-7}\cmidrule{8-10}
    & P  & R   &  F1 
    & P  & R   &  F1  
    & P  & R   &  F1
    \\
\midrule
MSCD-HAC& 39.0 & 91.0 & 54.6 & -- & -- & --   & -- & -- & -- \\
ALMSER-GB& 34.0 & 85.4 & 48.6 & 48.6 & 91.5 & 63.5 & -- & -- & -- \\
AutoFJ   & 52.3 & 50.0 & 51.1 & 30.3 & 23.4 & 26.4 & -- & -- & -- \\
PromptEM & 33.7 & 88.0 & 48.7 & 41.1 & \textbf{92.3} & 56.9 & 29.4 & \textbf{90.4} & 43.0 \\
Ditto & 24.0 & 76.6 & 36.5 & 48.8 & 91.5 & 63.6 & 40.9 & \underline{87.8} & 55.8 \\
MultiEM & 90.5 & 91.4 & 90.9 & 91.1 & 86.2 & 88.6 & 83.7 & 88.8 & 82.2 \\
\midrule
LLM4MEM(F) & 95.2 & 95.2 & 95.2 & \underline{91.7} & 89.4 & 90.6 & \underline{84.5} & 83.1 & 83.8 \\
LLM4MEM(L) & \underline{96.4} & \underline{96.4} & \underline{96.4} & 90.6 & 91.5 & \underline{91.1} & 84.3 & 86.7 & \underline{85.5} \\
LLM4MEM(Q) & \textbf{97.2} & \textbf{97.0} & \textbf{97.1} & \textbf{93.4} & \underline{92.0} & \textbf{92.8} & \textbf{86.6} & 85.8 & \textbf{86.2} \\
\bottomrule
\end{tabular}
}}
\end{table}

\begin{table}[!t]
    \footnotesize
    \centering
    \caption{Matching Performance of All the Methods. Best F1 in \textbf{bold}, second best in \underline{underline}. The symbol “--” denotes that the method can NOT produce any result after 7 days in our experimental settings. "MSCD-HAC", "ALMSER-GB" can NOT produce any result after 7 days in our experimental settings.\label{tab-f1-2}}
    \label{ablation}
    \setlength\tabcolsep{6.0pt}
    \renewcommand\arraystretch{0.7}
    {
    {
    \begin{tabular}{@{}lccccccccc@{}}
    \toprule
    \multirow{2.5}{*}{\bf Model} & \multicolumn{3}{c}{\bf Music-2M} & \multicolumn{3}{c}{\bf Person} &  \multicolumn{3}{c}{\bf Shopee}    \\
    \cmidrule(r){2-4}\cmidrule(r){5-7}\cmidrule{8-10}
    & P  & R   &  F1 
    & P  & R   &  F1  
    & P  & R   &  F1
    \\
\midrule
ALMSER-GB          & -- & -- & -- & -- & -- & -- & -- & -- & -- \\
MSCD-HAC           & -- & -- & -- & -- & -- & -- & -- & -- & -- \\
AutoFJ             & -- & -- & -- & -- & -- & -- & \textbf{45.9} & 24.2 & \underline{31.6} \\
PromptEM           & -- & -- & -- & -- & -- & -- & 2.2 & 6.6 & 3.3 \\
Ditto              & -- & -- & -- & -- & -- & -- & 3.4 & 10.0 & 5.1 \\
MultiEM            & 69.4 & 68.1 & 68.7 & 33.6 & \underline{39.9} & 36.5 & 34.5 & 21.1 & 26.2 \\
\midrule
LLM4MEM(F)  & \underline{71.4} & 71.4 & \underline{71.4} & 38.1 & 37.9 & 38.0 & 36.6 & 24.1 & 29.1 \\
LLM4MEM(L) & 69.4 & \underline{71.9} & 70.6 & \underline{38.8} & 39.8 & \underline{39.3} & {38.5} & \textbf{28.2} & \textbf{32.6} \\
LLM4MEM(Q)  & \textbf{73.5} & \textbf{73.2} & \textbf{73.3} & \textbf{41.6} & \textbf{42.0} & \textbf{41.8} & \underline{40.5} & \underline{25.1} & {31.0} \\
\bottomrule
\end{tabular}
}}
\end{table} 

We first assess LLM4MEM's matching performance against baselines, with results for 6 datasets reported in Tables~\ref{tab-f1-1} and Table~\ref{tab-f1-2}. 
We use LLM4MEM(F), LLM4MEM(L), and LLM4MEM(Q) for Falcon3-7B~\cite{Falcon3}, LLaMA3.1-8B~\cite{llama3}, and Qwen2.5-7B~\cite{qwen2.5}, respectively.
Compared to dual-table models like AutoFJ~\cite{Auto-FuzzyJoin}, Ditto~\cite{ditto}, and PromptEM~\cite{PromptEM}, LLM4MEM demonstrates superior performance on most datasets. 
PromptEM and Ditto achieved high recall on specific datasets (e.g., Music-20K, Music-200K) due to pre-trained language models but showed lower precision because our evaluation included all ground truth pairs as candidates.  
Other multi-table baselines, such as ALMSER-GB~\cite{ALMSER-GB} and MSCD-HAC~\cite{MSCD-HAC}, rely on complex computations, leading to poor efficiency. 
MSCD-HAC fails on most datasets, while ALMSER-GB struggles with large-scale data. MultiEM, a state-of-the-art method leveraging hierarchical merging and pre-trained models, achieves strong results across baselines.  
Overall, LLM4ME achieves state-of-the-art performance on most datasets, demonstrating effectiveness and robustness. Different LLMs perform consistently well, with Qwen2.5 excelling in multi-table tasks. The results confirm the importance of the MPLAC module.

\subsection{Ablation Study}
We evaluated the effectiveness of LLM4MEM's key modules by comparing it with variants, MultiEM \textit{w/o} MPLAC and MultiEM \textit{w/o} DPM, with results in Tables~\ref{tab-abstudy-1} and Table~\ref{tab-abstudy-2}.

LLM4MEM \textit{w/o} MPLAC excludes instruction fine-tuning and data regularization via LLMs. Results show the MPLAC module significantly improves dataset text quality, benefiting subsequent embedding. Without MPLAC, the F1 score dropped by 15.15\% compared to using Qwen2.5-7B, confirming its role in enhancing accuracy and robustness.  
LLM4MEM \textit{w/o} DPM uses only merging-stage predictions as final output. The pruning phase was found to generally improve performance, with F1 scores dropping by 1.6\% without DPM. This highlights the Density-aware Pruning module's ability to refine predictions for more accurate matching.

\begin{table}[h]
    \footnotesize
    \centering
    \caption{Ablation study on LLM4MEM. \textit{w/o} MPLAC refers to the experimental performance of removing the MPLAC module, \textit{w/o} DPM refers to removing the trimmed portion and using the model pre-matched tuple as the result.\label{tab-abstudy-1}}
    \label{ablation}
    \setlength\tabcolsep{6.0pt}
    \renewcommand\arraystretch{1.0}
    {
    {
    \begin{tabular}{@{}lccccccccc@{}}
    \toprule
    \multirow{2.5}{*}{\bf Model} & \multicolumn{3}{c}{\bf Geo} & \multicolumn{3}{c}{\bf Music-20K} &  \multicolumn{3}{c}{\bf Music-200K}    \\
    \cmidrule(r){2-4}\cmidrule(r){5-7}\cmidrule{8-10}
    & P  & R   &  F1 
    & P  & R   &  F1  
    & P  & R   &  F1
    \\
    \midrule
    LLM4MEM(Q)
    & \textbf{97.2}    & \textbf{97.0}  & \textbf{97.1}  
    & \textbf{93.4}    & \textbf{92.0}  & \textbf{92.8} 
    & \textbf{86.6}    & \textbf{85.8}  & \textbf{86.2} \\
    \cmidrule{2-10}
    \textit{w/o} MPLAC   
    & 72.4    & 71.6   & 72.0 
    & 82.0    & 82.9   & 82.4
    & 75.6    & 77.2   & 76.4  \\
    \textit{w/o} DPM   
    & 97.2  & 97.0  & 97.1   
    & 91.8   & 91.7  & 91.8    
    & 83.0  & 84.1  & 83.6 \\   
\bottomrule

\end{tabular}
}}
\end{table}

\begin{table}[!t]
    \footnotesize
    \centering
    \caption{Ablation study on LLM4MEM. \textit{w/o} MPLAC refers to the experimental performance of removing the MPLAC module, \textit{w/o} DPM refers to removing the trimmed portion and using the model pre-matched tuple as the result.\label{tab-abstudy-2}}
    \label{ablation}
    \setlength\tabcolsep{6.0pt}
    \renewcommand\arraystretch{1.0}
    {
    {
    \begin{tabular}{@{}lccccccccc@{}}
    \toprule
    \multirow{2.5}{*}{\bf Model} & \multicolumn{3}{c}{\bf Music-2M} & \multicolumn{3}{c}{\bf Person} &  \multicolumn{3}{c}{\bf Shopee}    \\
    \cmidrule(r){2-4}\cmidrule(r){5-7}\cmidrule{8-10}
    & P  & R   &  F1 
    & P  & R   &  F1  
    & P  & R   &  F1
    \\
    \midrule
    LLM4MEM(Q)
    & \textbf{73.5}    & \textbf{73.2}  & \textbf{73.3}  
    & \textbf{41.6}    & \textbf{42.0}  & \textbf{41.8} 
    & \textbf{40.5}    & \textbf{25.1}  & \textbf{31.0} \\
    \cmidrule{2-10}
    \textit{w/o} MPLAC   
    & 62.7    & 58.4   & 60.4 
    & 20.3    & 26.2   & 22.9
    & 22.1    & 13.4   & 16.7  \\
    \textit{w/o} DPM   
    & 69.1  & 71.0  & 70.1   
    & 39.2   & 40.0  & 39.6    
    & 39.3  & 24.9  & 30.5 \\   
\bottomrule

\end{tabular}
}}
\end{table}

\subsection{Hyperparameter Analysis}
\begin{figure}[h]
\includegraphics[width=\textwidth]{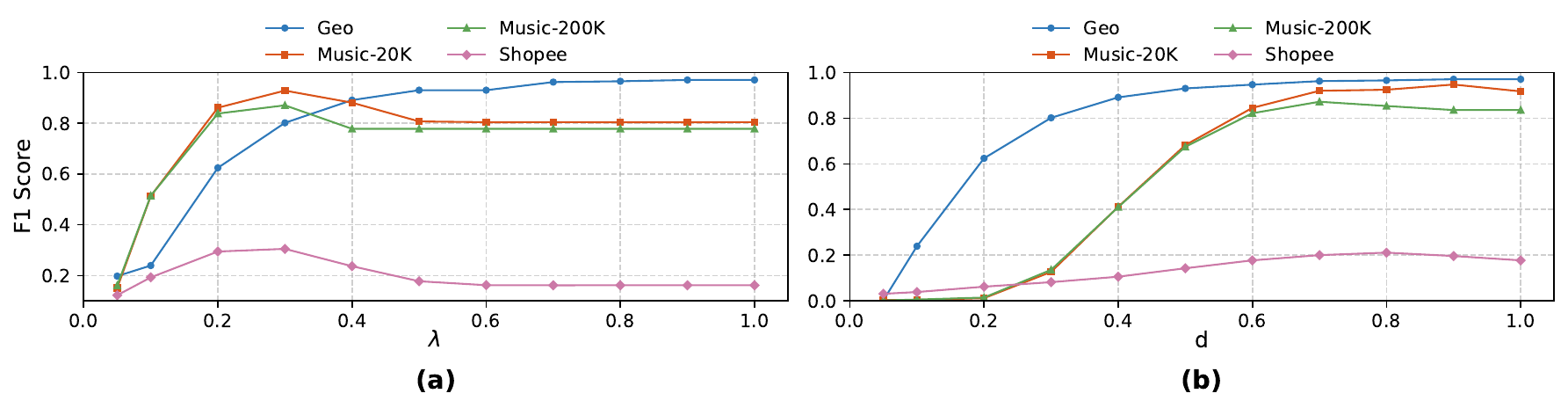}
\caption{The figure shows the sensitivity of key hyperparameters $\lambda$ (a) and $d$ (b) in the LLM4MEM method to experimental score F1.} \label{fig-v1}
\end{figure}

We further investigated the sensitivity of two key hyperparameters $\lambda$ and $d$ of the proposed LLM4MEM through the following experiments, as shown in detail in Figure~\ref{fig-v1} (a) and Figure~\ref{fig-v1} (b).

We conducted sensitivity analyses on the TOP1 similarity constraint parameter $\lambda$ (Section 4.2) and the density distance threshold $d$ (Section 4.3). 
For $\lambda$, results show it is highly sensitive near the median value (e.g., $\lambda=0.3$) across multiple datasets, with deviations from the optimal value causing significant F1 score drops. This sensitivity is particularly critical in datasets like Music-20K, Music-200K, and Shopee, where strict control of $\lambda$ is necessary.
For $d$, sensitivity manifests differently depending on the dataset. In low-value regions (e.g., Geo with $d<0.2$), smaller $d$ increases pseudo-anomalous entities, while in high-value regions (e.g., Music-20K with $d\in(0.6,0.9)$), larger $d$ risks misjudging core and reachable entities. Notably, extremely low $d$ values in Geo have the most pronounced impact on F1.
In conclusion, hyperparameters significantly influence experimental outcomes. The sensitivity of $\lambda$ is more universal and critical, especially in Music-20K, Music-200K, and Shopee, whereas $d$ requires dataset-specific adjustments based on data characteristics.

\section{Conclusion}

In this paper, we explore Multi-Table Entity Matching (MEM) to identify equivalent entities between multiple tables.
To unlock the power of Large Language Models (LLMs) for MEM, we propose a novel LLM-based framework for MEM, termed LLM4MEM.
First, the multi-style prompt-enhanced attribute coordination module resolves semantic inconsistencies across schemas using LLMs, eliminating manual alignment. 
Second, the transitive consensus embedding matching mechanism tackles the combinatorial explosion via bidirectional top-1 filtering and graph-based propagation. 
Third, the density-aware pruning module improves matching quality by removing noise through spatial constraints. 
Experiments on 6 datasets show LLM4MEM achieves state-of-the-art performance, with a 5.1\% F1 improvement over existing methods while maintaining linear complexity. 
Its unsupervised, schema-agnostic design ensures broad applicability. 
In future work, we will investigate domain adaptation and dynamic scenarios.

\subsubsection{\ackname}This work is supported by the National Natural Science Foundation of China (No.62406319), and the Postdoctoral Fellowship Program of CPSF (No.GZC20232968).

\bibliographystyle{splncs04}
\bibliography{mybibliography}

\end{document}